\documentclass[conference]{IEEEtran}

\usepackage{times}

\usepackage[numbers]{natbib}
\usepackage{multicol}
\usepackage[bookmarks=true]{hyperref}

\usepackage[utf8]{inputenc} 
\usepackage[T1]{fontenc}    
\usepackage{url}            
\usepackage{booktabs}       
\usepackage{amsfonts}       
\usepackage{nicefrac}       
\usepackage{microtype}      
\usepackage{xcolor}         

\usepackage{url}
\usepackage{graphicx}
\usepackage{caption} 
\usepackage{algorithm}
\usepackage[noend]{algpseudocode}
\algrenewcommand{\algorithmiccomment}[1]{\hfill// #1}

\usepackage{amsmath,amssymb}
\usepackage{multirow}
\usepackage{array,multirow,graphicx}
\usepackage{float}
\usepackage{wrapfig}
\usepackage{todonotes}
\usepackage{subcaption}
\usepackage{amsmath}

\graphicspath{{figures/}}

\newcommand{\bs}{\mathbf{s}}
\newcommand{\ba}{\mathbf{a}}
\newcommand{\br}{\mathbf{r}}
\newcommand{\bg}{\mathbf{g}}

\newcommand{\st}{\bs_t}
\newcommand{\at}{\ba_t}

\newcommand{\stp}{\bs_{t+1}}

\newcommand{\states}{\mathcal{S}}
\newcommand{\actions}{\mathcal{A}}

\newcommand{\qattn}{Q}

\newcommand{\obs}{\mathbf{o}}

\newcommand{\rgb}{\mathbf{b}}

\newcommand{\pcd}{\mathbf{p}}
\newcommand{\proprio}{\mathbf{z}}

\newcommand{\replay}{\mathcal{D}}

\DeclareMathOperator{\E}{\mathbb{E}}

\DeclareMathOperator*{\argmax}{arg\,max}

\newcommand{\goalpose}{\bg}

\newcommand{\real}{\mathbb{R}}
\newcommand{\combinedcpaths}{\ensuremath{\Xi}}
\newcommand{\cpath}{\ensuremath{\xi}}
\newcommand{\config}{\mathbf{c}}
\newcommand{\mpfN}{M}
\newcommand{\curvefN}{B}
\newcommand{\mpf}{m}
\newcommand{\curvef}{b}
\newcommand{\pathpi}{\pi}
\newcommand{\pathpip}{\mu}
\newcommand{\rankf}{Q^R}
\newcommand{\rankfp}{\sigma}

\newcommand{\methName}{Learned Path Ranking}
\newcommand{\methAcro}{LPR}

\newcommand{\bezierc}{B\'ezier}

\begin{document}

\title{Coarse-to-Fine Q-attention with\\Learned Path Ranking}

\author{\authorblockN{Stephen James and Pieter Abbeel}
\authorblockA{UC Berkeley\\
\{stepjam, pabbeel\}@berkeley.edu}
}

\maketitle

\begin{abstract}
We propose \methName\ (\methAcro), a method that accepts an end-effector goal pose, and learns to rank a set of goal-reaching paths generated from an array of path generating methods, including: path planning, \bezierc\ curve sampling, and a learned policy. The core idea being that each of the path generation modules will be useful in different tasks, or at different stages in a task. When \methAcro\ is added as an extension to C2F-ARM, our new system, C2F-ARM+\methAcro, retains the sample efficiency of its predecessor, while also being able to accomplish a larger set of tasks; in particular, tasks that require very specific motions (e.g. opening toilet seat) that need to be inferred from both demonstrations and exploration data. In addition to benchmarking our approach across 16 RLBench tasks, we also learn real-world tasks, tabula rasa, in 10-15 minutes, with only 3 demonstrations. Videos and code found at: \url{https://sites.google.com/view/q-attention-lpr}.
\end{abstract}

\IEEEpeerreviewmaketitle

\section{Introduction}
\label{sec:introduction}

\begin{figure}
\centering
\includegraphics[width=1.0\linewidth]{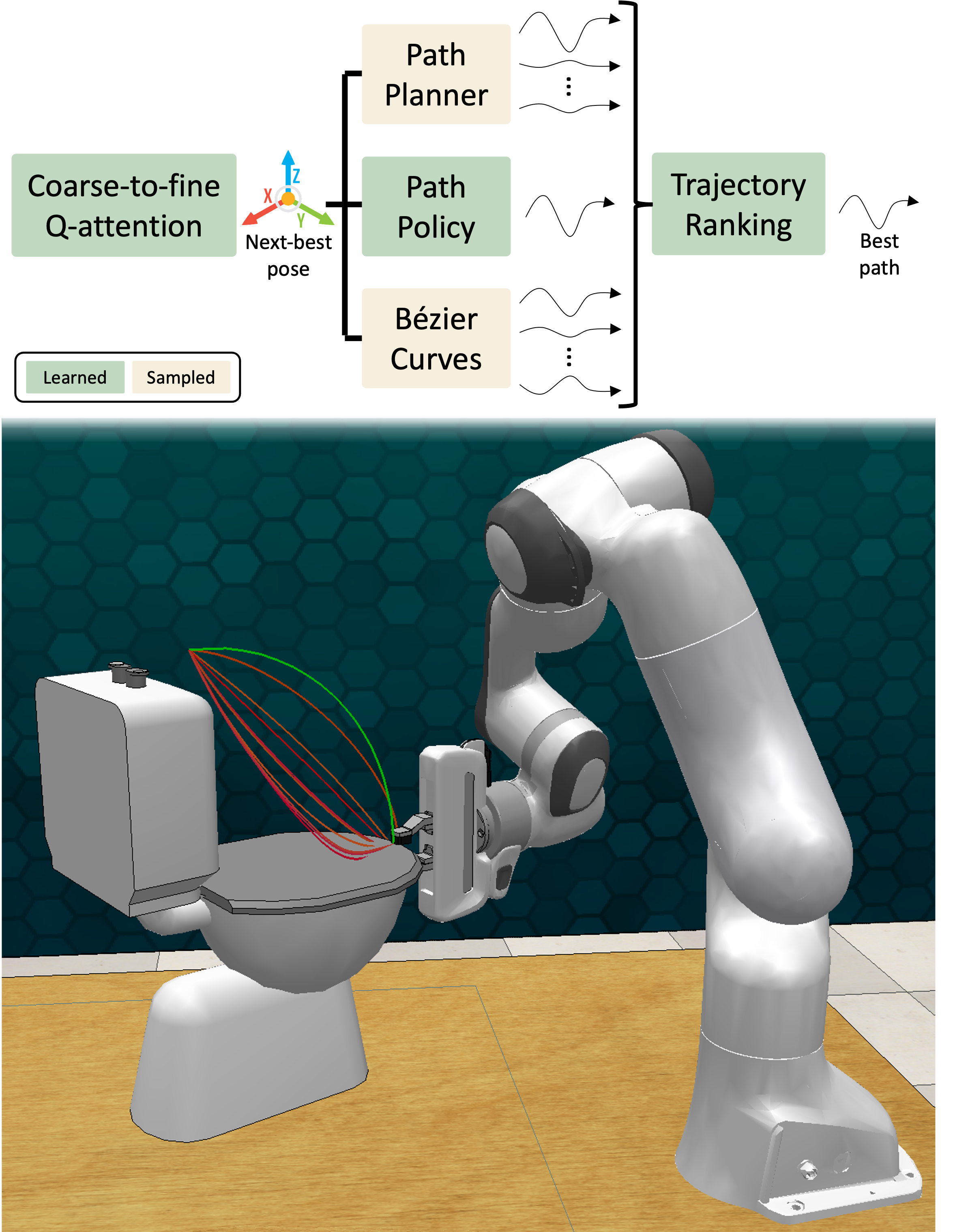}
\caption{C2F-ARM+\methAcro\ is an extension to C2F-ARM that, when given a next-best pose, learns to rank a set of paths generated from path planning, curve sampling, and a learned path policy. This allows for a larger set of tasks to be learned; in particular, tasks such lifting toilet seat, where only the green path (above) would result in a success.}
\label{fig:teaser}
\end{figure}

The search is ongoing for a general manipulation algorithm that operates on visual inputs and sparse rewards, yet is sufficiently sample-efficient to train in the real world. The recent works of Q-attention~\cite{james2022qattention} and coarse-to-fine Q-attention~\cite{james2021coarse} built into the Coarse-to-fine Attention-driven Robot Manipulation (C2F-ARM) system, have made training vision-based, sparse-rewarded reinforcement learning agents significantly easier. Using both a small number of demonstrations and exploration, C2F-ARM learns to output next-best poses which are then used by a motion planner to bring the robot to the predicted pose. This higher-level action space makes for highly-efficient learning in the real world. However, this comes at a cost: using motion planning means that the agent is unable to \textit{learn} to output very specific motions that may be needed for a task. An example of such a case is opening the toilet seat in Figure \ref{fig:teaser}, where a large number of motions are acceptable to get to the seat, but only a very limited number of motions are suitable to lift the seat open. Traditionally in these situations, hand-designed motions based on known kinematic constraints would be needed, e.g, a hard-coded linear path when a drawer handle is grasped, however, this requires prior knowledge about the objects in the scene. C2F-ARM assumes no prior knowledge of objects in the scene, and so a novel path-generation solution is required in order to allow C2F-ARM to accomplish tasks that need specific motions, without requiring explicit knowledge of kinematic constraints of objects.

Our solution is simple, yet highly effective. Given the next-best pose from the coarse-to-fine Q-attention, we sample a large number of collision-free and in-collision planned paths and curves; this is in contrast to C2F-ARM, which uses path planning to generate a single path. With these multiple paths, we leverage a learned ranking Q-function to choose the highest valued path, which is then executed on the robot. The intuition is that a subset of these paths will be desirable, and the algorithm can learn to choose which paths are appropriate in certain situations. In addition to this, we also learn a behavior cloning policy that is only trained on paths (from path planning and curves sampling) that \textit{eventually} lead to successfully completing the task. This has the effect of capturing the preferences for the paths that are useful for solving the task, and then distilling those preferences into the policy. Generated paths from the policy are also ranked, meaning that they are not run on the robot until they are deemed better than the sampled alternatives. 

\begin{figure}
\centering
\includegraphics[width=1.0\linewidth]{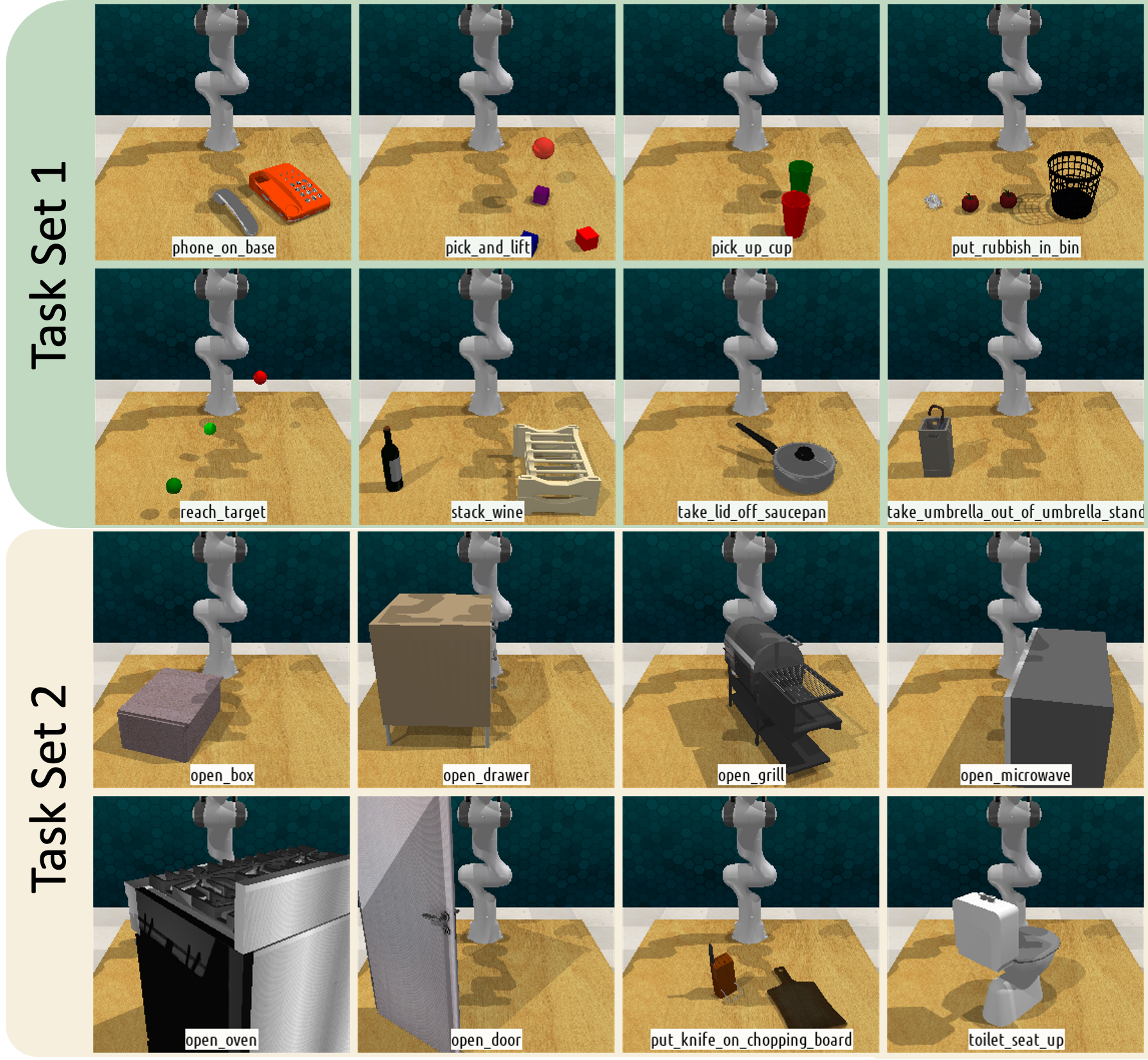}
\caption{In simulation, our method is evaluated on a total of 16 RLBench~\cite{james2019rlbench} tasks. \textit{Task Set 1} are tasks that have previously been shown to perform well with C2F-ARM. \textit{Task Set 2} are tasks that were chosen due to requiring very particular motion that are necessary for task completion (e.g. opening or closing objects). Note that the position and orientation of objects are placed randomly at the beginning of each episode.}
\label{fig:sim_taskset}
\end{figure}

We benchmark our method, C2F-ARM+\methAcro\, on 16 simulated RLBench~\cite{james2019rlbench} tasks (Figure \ref{fig:sim_taskset}) and compare against C2F-ARM (without \methAcro). We show that C2F-ARM+\methAcro\ can accomplish a wider range of tasks, and much like C2F-ARM, is capable of learning sparsely-rewarded real-world tasks from only 3 demonstrations.

To summarize, our main contributions are three fold: \textbf{(1)} \methName\ (\methAcro), a Q-function that learns to rank paths generated from different sources, including path planners, \bezierc\ curve sampling, or even a learned path policy; \textbf{(2)} A hybrid set of path planning modules that combines both sample-based planning and learned planning, allowing for the system to rely less on sampling as training progresses; \textbf{(3)} A new system, C2F-ARM+\methAcro, which improves the performance over C2F-ARM on a large set of RLBench tasks.

\section{Related Work}
\label{sec:related_work}

Recent trends in learning for manipulation have seen reinforcement learning (RL) algorithms trained on a variety of tasks, including lego stacking~\cite{haarnoja2018composable}, block lifting~\cite{james20163d}, cloth manipulation~\cite{matas2018sim}, pushing~\cite{pinto2017asymmetric}, and grasping~\cite{kalashnikov2018qt, james2019sim}, to name but a few. In this work, we extend C2F-ARM~\cite{james2021coarse} --- an algorithm that discretizes the action space in a coarse-to-fine manner for efficient learning via a Q-attention network. The predecessor system, ARM~\cite{james2022qattention}, has recently been extended to a Bingham policy parameterization~\cite{james2022bingham} to improve training stability. In this work, our method does not directly extend coarse-to-fine Q-attention, but rather looks to improve how to use the next-best pose output through the use of path ranking.

\subsection*{Learning in path planning}

Recent work has attempted to increase the speed of planning by predicting collisions for object poses from scene and query object point clouds~\cite{danielczuk2021object}. \textit{Mitrano et al.}~\cite{mitrano2021learning} learned a function to rule out edges in a search tree for manipulating cables. Rather than attempting to modify existing planners with learned components, we instead propose to filter the output of path planners (and other path generation techniques) via a learned ranking Q-function. A filtering stage is common for path planning, i.e. generating many candidate paths and then choosing the best based on predefined metrics (e.g. path length, joint constraints, etc). However, to the best of our knowledge, this is the first work to propose to \textit{learn} this filtering from sparse rewards via RL. 

The path ranking function mentioned above, learns to rank a set of paths generated from path planning, curve sampling, and a learned path policy. For the latter, we use behavior cloning to learn to predict paths given a start and goal configuration. A modest body of work has explored using path planners to generate training labels to train neural networks that iteratively predict paths that bring the agents closer to target state~\cite{tamar2016value,srinivas2018universal,qureshi2019motion,bency2019neural,bhardwaj2020differentiable}. Work has also explored learning to regress to paths in an unsupervised manner, without relying on generating paths from a planner~\cite{pandy2020unsupervised}. Combining RL and path planning has been explored in PRM-RL~\cite{faust2018prm}, where an RL agent is trained to determine connectivity, while PRM is used for global planning.

We would like to briefly mention that a loose connection could be made between our work and the paradigm of learning from human preference~\cite{furnkranz2010preference}, however we do not learn a reward function, nor do we ever ask the user to compare two paths or trajectories.

\section{Background}
\label{sec:background}

\subsection{Reinforcement Learning}

The reinforcement learning paradigm assumes an agent interacting with an environment consisting of states $\bs \in \states$, actions $\ba \in \actions$, and a reward function $R(\st,\at)$, where $\st$ and $\at$ are the state and action at time step $t$ respectively. The goal of the agent is then to discover a policy $\pi$ that results in maximizing the expectation of the sum of discounted rewards: $\E_\pi [\sum_t \gamma^t R(\st, \at)]$, where future rewards are weighted with respect to the discount factor $\gamma \in [0, 1)$. Each policy $\pi$ has a corresponding value function $Q(s, a)$, which represents the expected return when following the policy after taking action $\ba$ in state $\bs$.

The Q-attention module~\cite{james2022qattention} (discussed in Section \ref{sec:back:arm}) builds on top of Deep Q-learning \cite{mnih2015human}; a method that approximates the value function $Q_\psi$, with a deep convolutional network, whose parameters $\psi$ are optimized by sampling mini-batches from a replay buffer $\replay$ and using stochastic gradient descent to minimize the loss: $\E_{(\st, \at, \stp) \sim \replay} [ (\br + \gamma \max_{\ba'}Q_{\psi'}(\stp, \ba') - Q_{\psi}(\st, \at))^2]$, where $Q_{\psi'}$ is a target network --- a periodic copy of the online network $Q_\psi$.

\subsection{Coarse-to-fine Attention-driven Robot Manipulation (C2F-ARM)}
\label{sec:back:arm}

C2F-ARM~\cite{james2021coarse} is learning-based manipulation system that can solve sparsely-rewarded, image-based tasks. It can be summarized by 2 core phases. \textbf{Phase 1} consists of the coarse-to-fine Q-attention agent, that takes a voxelized scene and learns what part of it to `zoom' into. When this `zooming' behavior is applied iteratively, it results in a near-lossless discretization of the translation space, and allows the use of a discrete action, deep Q-learning method. \textbf{Phase 2} is a control agent that accepts the predicted next-best pose and executes a series of actions to reach the given goal pose. In C2F-ARM~\cite{james2021coarse}, this control agent was a motion planner, though in practice could also be a reinforcement learning agent. 

To overcome the exploration challenge imposed by the sparse reward setup, C2F-ARM makes use of a small number of demonstrations. Rather than simply inserting these demonstrations directly into the replay buffer, the system uses a \textit{keyframe discovery} method that aids the Q-attention network to quickly converge and suggest meaningful points of interest. C2F-ARM also makes use of \textit{demo augmentation}, which increases the amount of initial demo transitions in the replay buffer.

As our method does not directly extend Q-attention, but rather looks to improve how to use the next-best pose output, we refer the reader to \textit{James et al.}~\cite{james2021coarse} for a detailed summary of coarse-to-fine Q-attention.

\section{Method}
\label{sec:method}

\begin{algorithm}[tb]
\caption{C2F-ARM+\methAcro}
\label{alg:arm}
\begin{algorithmic}[1]
    \State Initialize Coarse-to-fine Q-attention $\qattn$.
    \State Initialize $\pathpi_\pathpip, \rankf_\rankfp$ with parameters $\pathpip, \rankfp$.
    \State Initialize buffer $\replay$ with demos; apply \textbf{keyframe discovery} and \textbf{demo augmentation}.
    \For{each iteration}
	    \For{each environment step $t$}
	        \State $\obs_t \leftarrow (\rgb_t, \pcd_t, \proprio_t)$
	        \State $\goalpose_t \leftarrow \qattn(\obs_t)$ \Comment{Get next-best pose}
	        
	        \State $[\cpath_{1}^{\mpf}, \ldots, \cpath_{\mpfN}^{\mpf}] \leftarrow \mpf(\proprio_t, \goalpose_t, \mpfN)$   \Comment{Path planning}
	        \State $[\cpath_{1}^{\curvef}, \ldots, \cpath_{\curvefN}^{\curvef}] \leftarrow \curvef(\proprio_t, \goalpose_t, \curvefN)$   \Comment{\bezierc\ sampling}
	        \State $\cpath^{\pathpi} = \pathpi(\obs_t, \goalpose)$ if valid else $\emptyset$ \Comment{Path policy}
	        
	        \State $\combinedcpaths = [\cpath_{1}^{\mpf}, \ldots, \cpath_{\mpfN}^{\mpf}, \cpath_{1}^{\curvef}, \ldots, \cpath_{\curvefN}^{\curvef}, \cpath^{\pathpi}]$
	        \State $\cpath^* = \argmax_\cpath \rankf_{\rankfp}(\combinedcpaths)$
	        
    	    \State $\obs_{t+1}, \br \leftarrow env.step(\cpath^*)$  \Comment{Run trajectory on $\cpath^*$}
            \State $\mathcal{D} \leftarrow \replay \cup \left\{(\obs_t, \goalpose_t, \br, \obs_{t+1}, \cpath^*)\right\}$ \Comment{Store transition}
	        
	    \EndFor
    	\For{each gradient step}
    	    \State Update $\qattn$
    	    \State $\pathpip \leftarrow \pathpip - \lambda_{\pathpi} \hat \nabla_{\pathpip} J_{\pathpi}(\pathpip)$ \Comment{Update parameters}
    	    \State $\rankfp \leftarrow \rankfp - \lambda_{\rankf} \hat \nabla_{\rankfp} J_{\rankf}(\rankfp)$ \Comment{Update parameters}
    	    \State $\rankfp' \leftarrow \tau \rankfp + (1-\tau) \rankfp'$ \Comment{Update target network}
    	\EndFor
    \EndFor
\end{algorithmic}
\end{algorithm}

Our \methName\ (\methAcro) consists of 3 path generating modules: path planning, \bezierc\ curve sampling, and a learned path policy, which are then ranked using a learned ranking Q-function. The C2F-ARM system assumes we are operating in a partially observable Markov decision process (POMDP), where an observation $\obs$ consists of an RGB image, $\rgb$, an organized point cloud, $\pcd$, and proprioceptive data, $\proprio$. Actions consist of a 6D (next-best) pose and gripper action, and the reward function is sparse, giving $1$ on task completion, and $0$ for all other transitions. We define a path of length $T$ to consist of a series of robot joint configurations: $\cpath = [\config_1, \ldots, \config_T]$, where $\config \in \real^d$ for a $d$ DoF arm.

\begin{figure*}
\centering
\includegraphics[width=1.0\linewidth]{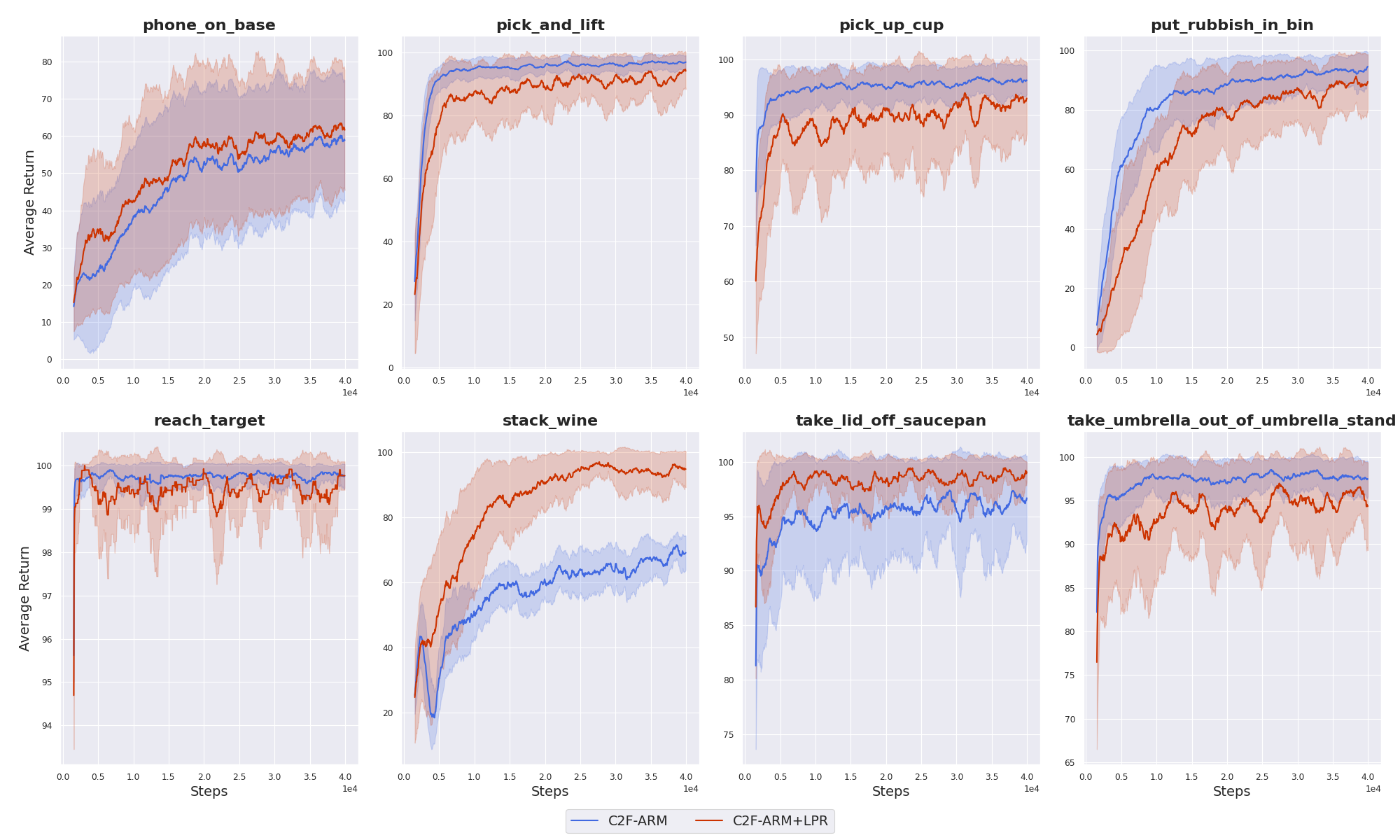}
\caption{Learning curves for 8 RLBench tasks. Both methods only receive 10 demos, which are stored in the replay buffer prior to training. Note, the \textbf{goal here is not to outperform C2F-ARM on these tasks}, but to instead show that there is no loss in performance. In Figure \ref{fig:c2fvspath_new_tasks} however, the goal is to outperform C2F-ARM, as those tasks require particular motions. Solid lines represent the average evaluation over 5 seeds, while the shaded regions represent the $std$.}
\label{fig:c2fvspath_orig_tasks}
\end{figure*}

\subsection{Path Sampling}

Two of the path generation modules involve sampling from a motion planning function $\mpf(\obs, \proprio, \goalpose)$ and a \bezierc\ curve function $\curvef(\obs, \proprio, \goalpose)$, where $\obs$ contains the point-cloud data used for collision checking, $\proprio$ is the current proprioceptive data (containing the start pose), and where $\goalpose$ is the next-best pose output from the Q-attention. For the path planning function $\mpf$, $\mpfN$ paths are generated ($\mpfN=20$ in our case) by first sampling $\mpfN$ joint configurations that satisfy the goal pose from the next-best pose agent, and then running RRTConnect~\cite{kuffner2000rrt} from the Motion Planning Library (OMPL)~\cite{sucan2012ompl} to produce a set of paths $[\cpath_{1}^{\mpf}, \ldots, \cpath_{\mpfN}^{\mpf}]$. When running planning, a portion of the paths will have collision checking disabled; this is to allow for tasks that require some amount of contact, for example pushing objects. Each generated path will also have a boolean flag representing if the path collides with the scene. 

The \bezierc\ curve function $\curvef$ is the other path generation module. Before motivating this second module, we first recall the definition a \bezierc\ curve. Given $n+1$ control points $(P_0, \ldots, P_n)$, a \bezierc\ curve is defined as:
\begin{equation}
P(t) = \sum^{n}_{i=0} B^n_i(t) \cdot P_i, \qquad t \in [0, 1],
\end{equation}
where t is a point along the path, and $B(t)$ is the Bernstein polynomial:
\begin{equation}
B^n_i(t) = {n \choose i} t^i(1-t)^{n-i}.
\end{equation}

For the curve function $\curvef$, $\curvefN$ Cartesian paths are generated ($\curvefN=20$ in our case) by first sampling $\curvefN$ Quadratic \bezierc\ curves (\bezierc\ curves with 3 control points), where the first and last control points are fixed at the start and end positions of the end-effector, and the middle control point is sampled from a Gaussian distribution, with mean set to be between the start and end control points, and the variance set to $0.2$. Using inverse kinematics, we convert these curves to robot configurations to give $[\cpath_{1}^{\curvef}, \ldots, \cpath_{\curvefN}^{\curvef}]$.

Whereas sample-based path planning occasionally produces smooth paths, \bezierc\ curves are almost always smooth, especially when $n$ is low, which make them ideal for tasks that require opening and closing objects; however they do not factor in collisions when being generated. The core idea of using both, followed by ranking, is that each of the path generation modules will be useful in different tasks, or even for the same tasks but at different stages in the task.

\subsection{Path Learning}
\label{sec:path_learning}

\begin{figure*}
\centering
\includegraphics[width=1.0\linewidth]{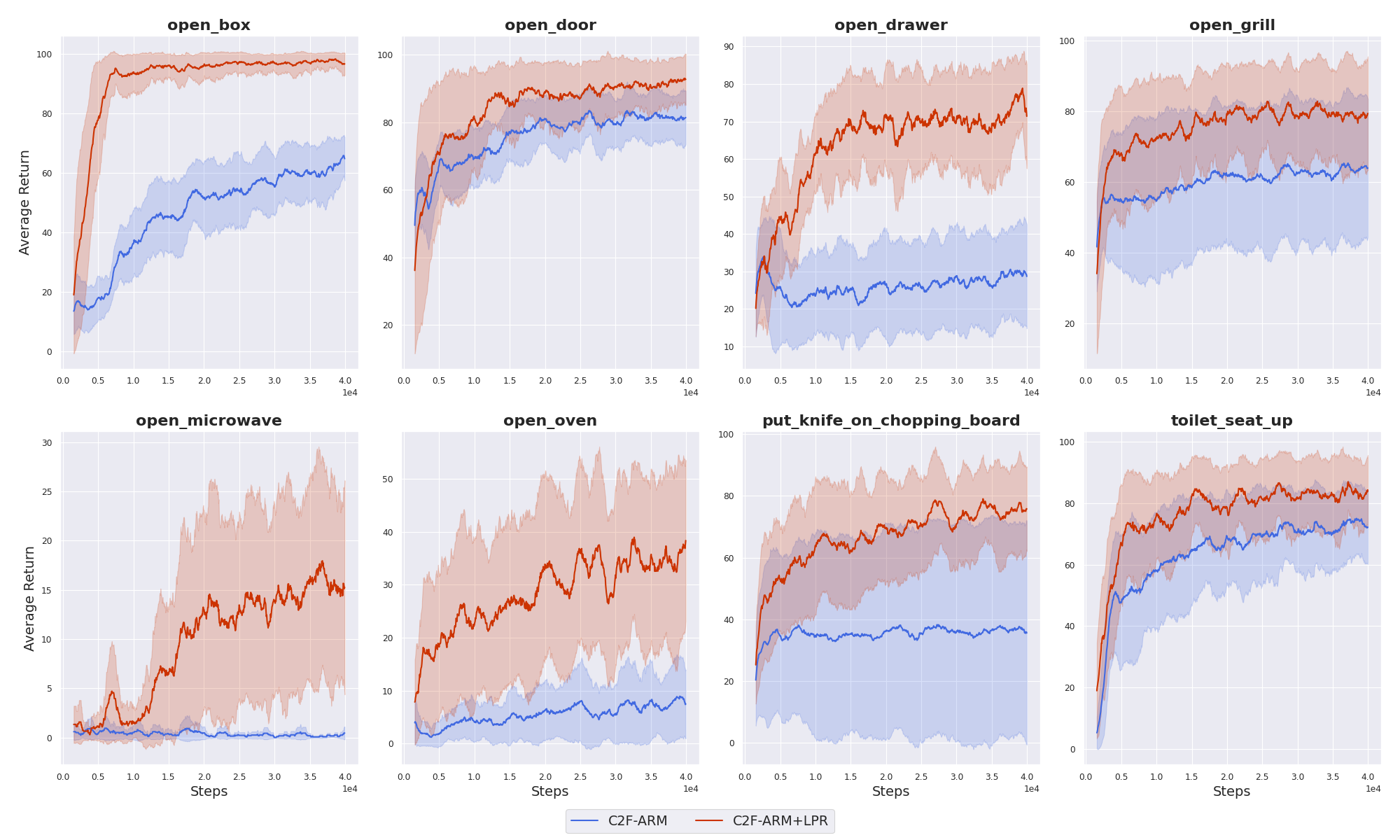}
\caption{Learning curves for an additional 8 RLBench tasks. Both methods only receive 10 demos, which are stored in the replay buffer prior to training. Unlike Figure \ref{fig:c2fvspath_orig_tasks}, the \textbf{goal here is to outperform C2F-ARM}, as these tasks require particular motions; e.g. opening doors, drawers, etc. Solid lines represent the average evaluation over 5 seeds, while the shaded regions represent the $std$.}
\label{fig:c2fvspath_new_tasks}
\end{figure*}

As we will see in the results section, outputting and ranking paths from both path planning and curve sampling can already improve performance when compared to C2F-ARM. However, even if both $\mpfN$ and $\curvefN$ are sufficiently large, it still may be the case that none of the candidate paths are suitable for the given situation. Therefore, rather than relying solely on sampling, it motivates the need to learn to predict suitable paths. Our 3rd path generation module takes the form of a behavior cloning path policy $\cpath^{\pathpi} = \pathpi(\proprio, \goalpose)$, that is trained on paths that eventually lead to the episode succeeding; this includes being trained on demonstrations that have already been stored in the replay buffer. 

Note that depending on the role that we want the path policy to play, it does not \textit{necessarily} need access to vision input. In this paper, we look to have the path policy complement the other 2 modules, rather than eventually replace them. For the path policy to be able to replace the other two modules, it would indeed need access to vision. We mention this further when discussing future work (Section \ref{sec:conclusion}). In summary, there is a cost-benefit trade-off to be made: our design decision makes the policy extremely lightweight. Adding vision to the path policy is not computationally trivial: recall that C2F-ARM, by design, is agnostic to the number of cameras in the scene, therefore, to include vision in our path policy, we would need to take a similar approach as C2F Q-attention, which fuses all cameras to a conical voxel grid, and thus would dramatically increase the computation cost of our system. We discuss avenues to reduce this computational overhead in Section \ref{sec:conclusion}.

The policy is represented as a fully-connected network (with output sizes 64, 256, 256, $d$), where $d$ is the DoF of the arm ($7$ in our case), and where each layer is followed by ReLU activations, except for the final which uses a linear activation. We use a mean squared error loss to regress the path policy to valid paths; i.e. $J_{\pathpi}(\pathpip) = \sum(\cpath^{\pathpi} - \cpath^{*})^2$, where $\cpath^{*}$ is a path that previously resulted in success.

While the agent is exploring during training, the path $\cpath^{\pathpi}$ is not directly given to the learned path ranking function, but is first analyzed to see if it is valid. The path is marked valid if its distance in configuration space is less than the mean of the paths from both the path planning and curve sampling modules. The intuition here is that at the beginning of training, the predicted configurations will be undesirable (e.g. configurations that would not lead to smooth motions) because it has not been trained on a sufficient number of successful paths; and so filtering based on length allows for invalid paths not to take part in the ranking. This is important, because paths at the beginning of training will be heavily out of distribution, increasing the chance that the Q-function may (mistakenly) assign a high value (due to the Q-value overestimation phenomenon). As the agent collects more exploration data, and the path policy becomes better at identifying the subtleties of the paths needed to accomplish the given task, the algorithm can become less dependent on the sample-based modules, and more dependent on its own predicted paths, which are more likely to be close to the path needed to complete a task.

\subsection{Learned Path Ranking}
Combining the output from the 3 modules above, we now have a set of paths $\combinedcpaths = [\cpath_{1}^{\mpf}, \ldots, \cpath_{\mpfN}^{\mpf}, \cpath_{1}^{\curvef}, \ldots, \cpath_{\curvefN}^{\curvef}, \cpath^{\pathpi}]$. Note that as discussed in the previous section, $\cpath^{\pathpi}$ is excluded from $\combinedcpaths$ if it is invalid. The goal now is to learn a ranking Q-function $\rankf(\cpath)$ with parameters $\rankfp$, which are optimized by sampling mini-batches from a replay buffer $\replay$ and using stochastic gradient descent to minimize the loss: $J_{\rankf}(\rankfp) = \E_{(\cpath_{t}, \cpath_{t+1}) \sim \replay} [ (\br + \gamma \rankf_{\rankfp'}(\cpath_{t+1}) - \rankf_{\rankfp}(\cpath))^2]$, where $\rankf_{\rankfp'}$ is a target network; a periodic copy of the online network $\rankf_{\rankfp}$. The reward here is the same sparse-reward used for the Q-attention; i.e. the ranking Q-function operates in the same underlying POMDP.

\begin{figure*}
\centering
\includegraphics[width=1.0\linewidth]{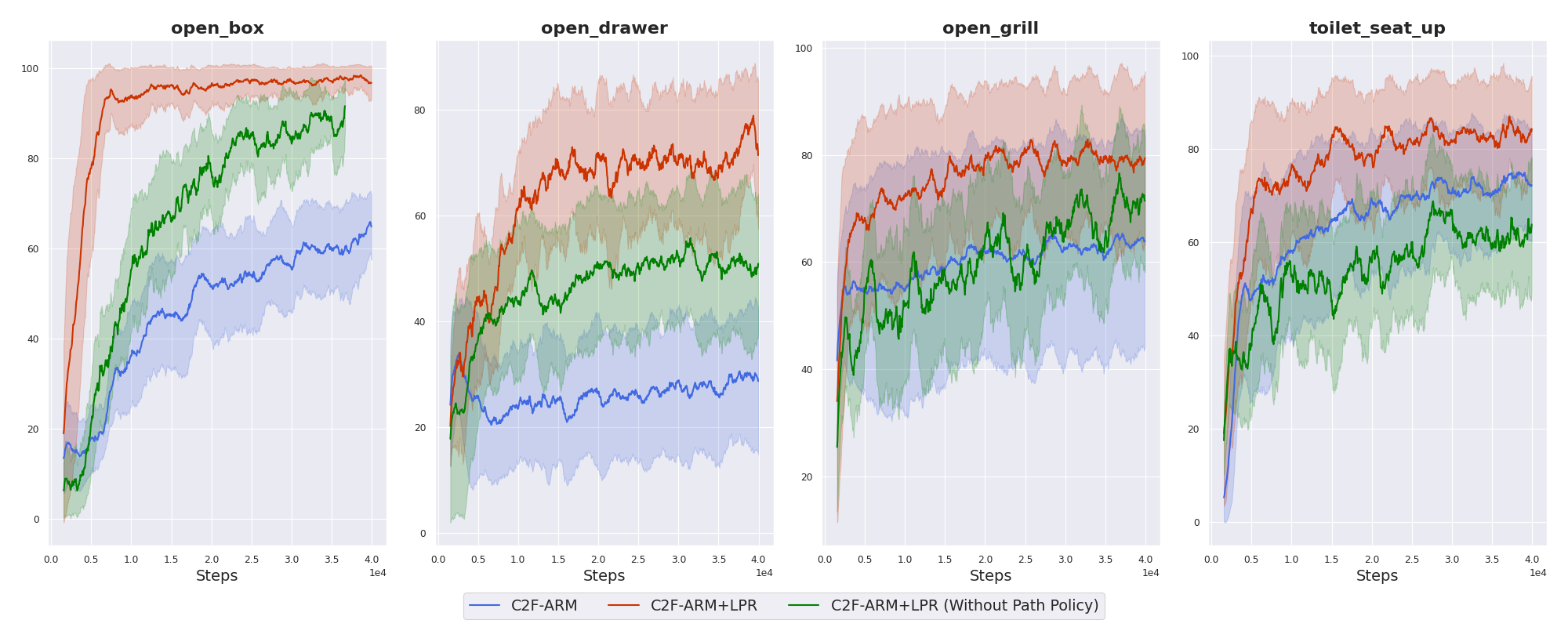}
\caption{Ablation of the learned path policy on a set of 4 RLBench tasks. All methods only receive 10 demos, which are stored in the replay buffer prior to training. Solid lines represent the average evaluation over 5 seeds, while the shaded regions represent the $std$.}
\label{fig:c2fvspathvsLT}
\end{figure*}

\begin{figure}
\centering
\includegraphics[width=1.0\linewidth]{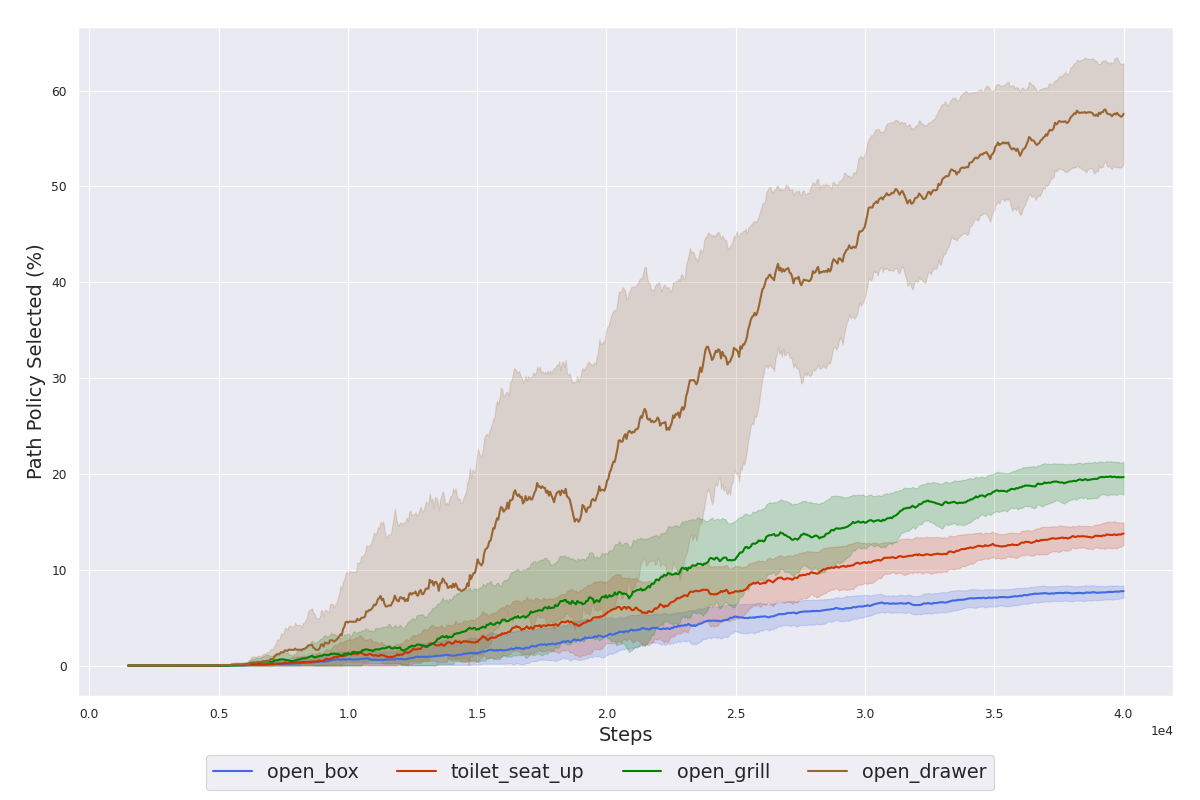}
\caption{How often the path ranking function predicts the learned policy trajectory to be the highest valued trajectory as training progresses. Solid lines represent the average evaluation over 5 seeds, while the shaded regions represent the $std$.}
\label{fig:percentage_policy_used}
\end{figure}

The architecture for the path policy is composed of shared fully-connected layers (with output sizes 64, 128, 1024) operating on each of the configurations in a path, followed by a max-pooling layer across configuration features, and finally processed by a further three fully-connected layers (with output sizes 512, 512, 1) , where the output is a Q-value. All layers use ReLU activations, except for the final which uses a linear activation.

\subsection{Runtime cost}

Despite the addition of the 3 path generation modules along with the learned path ranking, the runtime cost only increases by a factor of $1.2$. This increase in runtime is small for two primary reasons: \textbf{(1)} The two additional networks (path policy and path ranking Q-function) are small fully-connected layers, that are cheap for both forward passes and gradient updates, equating to a negligible runtime increase in comparison to the more expensive Q-attention runtime; \textbf{(2)} The planning within C2F-ARM would generate several candidate joint configurations, along with several path plans for each of these configurations, and choose the one with the shortest path; this is similar to our approach, but rather than choosing the path based on some heuristic (e.g. shortest path), we instead allow the ranking function to choose the best path. 

\section{Results}
\label{sec:simulation_results}

In the result sections, we show the following 4 core takeaways:

\begin{enumerate}
    \item We are able to maintain the high performance of C2F-ARM on tasks where learning particular motions is not necessary.
    \item On tasks where learning particular motions is crucial, our method outperforms C2F-ARM.
    \item Our proposition to have a learned policy that outputs paths does contribute to the success of the method.
    \item Like C2F-ARM, our real-world results show that the sample-efficiency in simulation is also present when training from scratch in the real world.
\end{enumerate}

\subsection{Quantitative Simulation Results}
\label{sec:simulation_results}

\begin{figure*}
\centering
\includegraphics[width=1.0\linewidth]{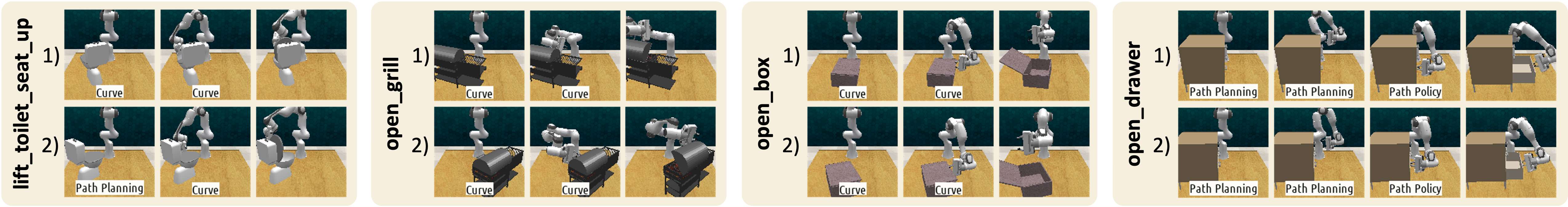}
\caption{Qualitative simulation results. We visualize two episodes for 4 trained C2F-ARM+\methAcro\ agents across 4 tasks. At each step, we annotate the frame with the path generation method that was ranked the highest by the \methAcro.}
\label{fig:episode_breakdown}
\end{figure*}

For our simulation experiments, we use RLBench~\cite{james2019rlbench}, which emphasizes vision-based manipulation and gives access to a wide variety of tasks with demonstrations. Each task has a  sparse reward of $1$, which is given only on task completion, and $0$ otherwise. We follow \textit{James et al.}~\cite{james2021coarse}, where we use a coarse-to-fine depth of $2$, each with a voxel grid size of $16^3$.

To compare C2F-ARM with and without \methAcro, we select the same 8 tasks as in previous work~\cite{james2022qattention, james2021coarse}; these are tasks that are achievable from using only the front-facing camera. These 8 tasks are also ones where learning particular motions is not necessary, and so in Figure \ref{fig:c2fvspath_orig_tasks}, the goal is to show that there is no loss in performance when using \methAcro. The results illustrate that C2F-ARM+\methAcro\ achieves either similar or better performance than C2F-ARM. This is unsurprising, as all of these tasks can be completed using a range of motions; i.e, they are all tasks that mostly follow a pick-and-place style.

Next, in Figure \ref{fig:c2fvspath_new_tasks}, we show that C2F-ARM+\methAcro\ can improve performance on tasks where only a small subset of all motions are suitable; these include tasks such as opening and closing objects, and tasks that require a sliding motion. The figure shows that across 8 such tasks, C2F-ARM+\methAcro\ attains similar or better performance. 

For our final set of simulation experiments, we aim to understand the effect of the learned path policy. For this, we choose a set of 4 tasks that operate on 4 different objects: box, drawer, grill, and toilet, where the task is to open each of them. Figure \ref{fig:c2fvspathvsLT} shows that there is indeed a benefit to including the learned path policy in the ranking process. Related to this experiment, in Figure \ref{fig:percentage_policy_used} we investigate how often the path predicted by the path policy is chosen by the ranking function. Interestingly, when comparing both Figures \ref{fig:c2fvspathvsLT} and \ref{fig:percentage_policy_used}, the percentage of times the policy path is chosen does not need to be high for it to have a positive effect on the success of a task. We hypothesize, that when $\mpfN$ and $\curvefN$ is sufficiently large, the change of a smooth, suitable path is high, rendering the learned path inferior. However, in rare cases when no suitable paths are generated, the path predicted by the policy may be the only suitable path.

\begin{figure}
\centering
\includegraphics[width=1.0\linewidth]{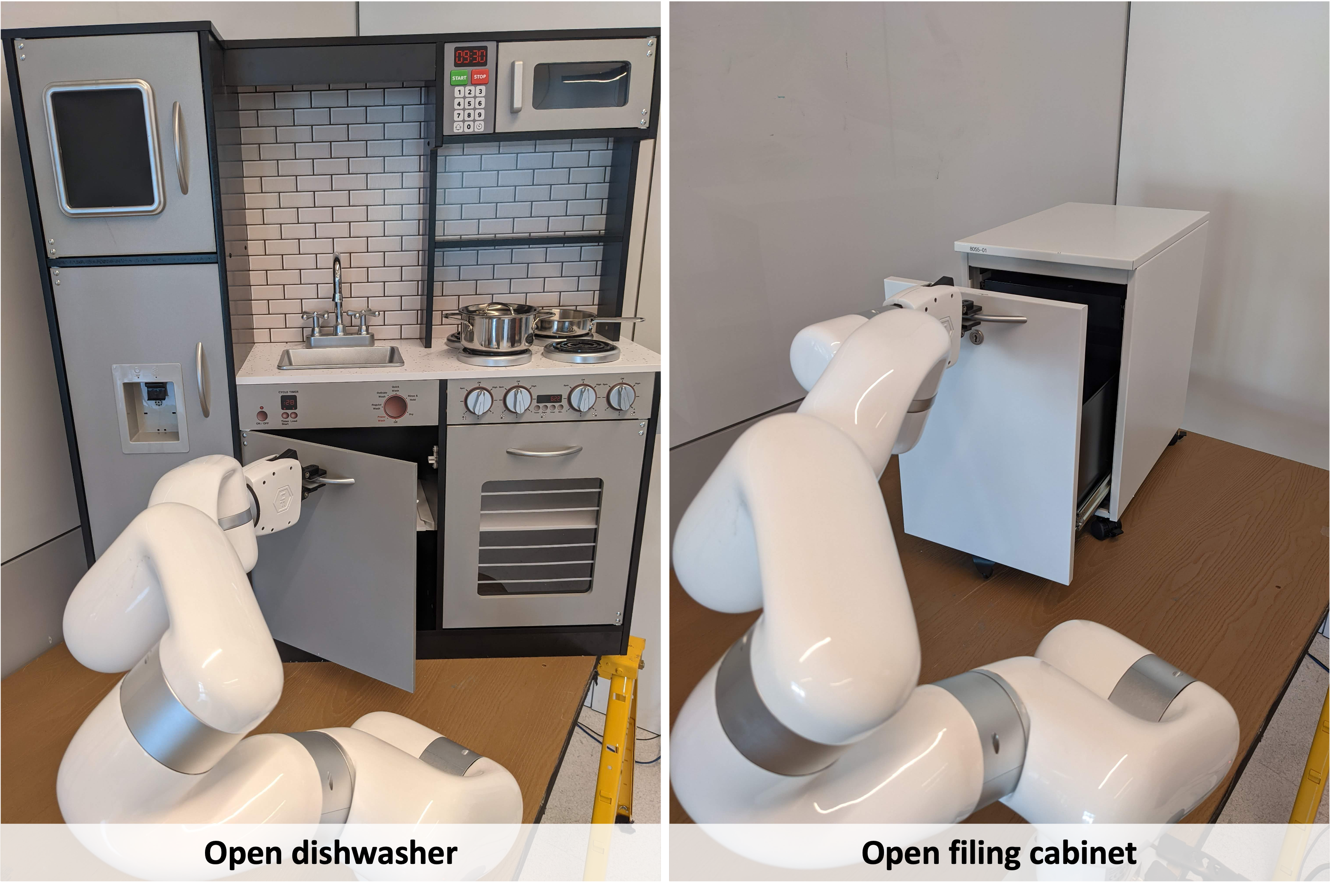}
\caption{The two real-world tasks for our qualitative evaluation; (left) open toy dishwasher, (right) open filing cabinet.}
\label{fig:real_exp}
\end{figure}

\subsection{Qualitative Simulation Results}
\label{sec:qual_sim_results}

To get a better understanding of how the path ranking behaves across tasks, we perform a qualitative analysis in Figure \ref{fig:episode_breakdown}, where we visualize episodes for 4 trained C2F-ARM+\methAcro\ agents across 4 tasks. At each step, the frames are annotated with the path generation method that was ranked the highest. The results match up with Figure \ref{fig:percentage_policy_used}, where the \textit{open\_drawer} task uses the path policy more than the other three tasks. Having the path policy ranked highest for the sliding motion of the drawer is intuitive, as both a curved path or collision-free path planned path does not seem appropriate. However, for the other 3 tasks, each of the objects have revolute joints (\textit{lift\_toilet\_seat\_up}, \textit{open\_box}, \textit{open\_grill}), making curved paths a natural choice to correctly articulating the objects.

\subsection{Qualitative Real-world Results}
\label{sec:real_world_results}

\begin{figure}
\centering
\includegraphics[width=1.0\linewidth]{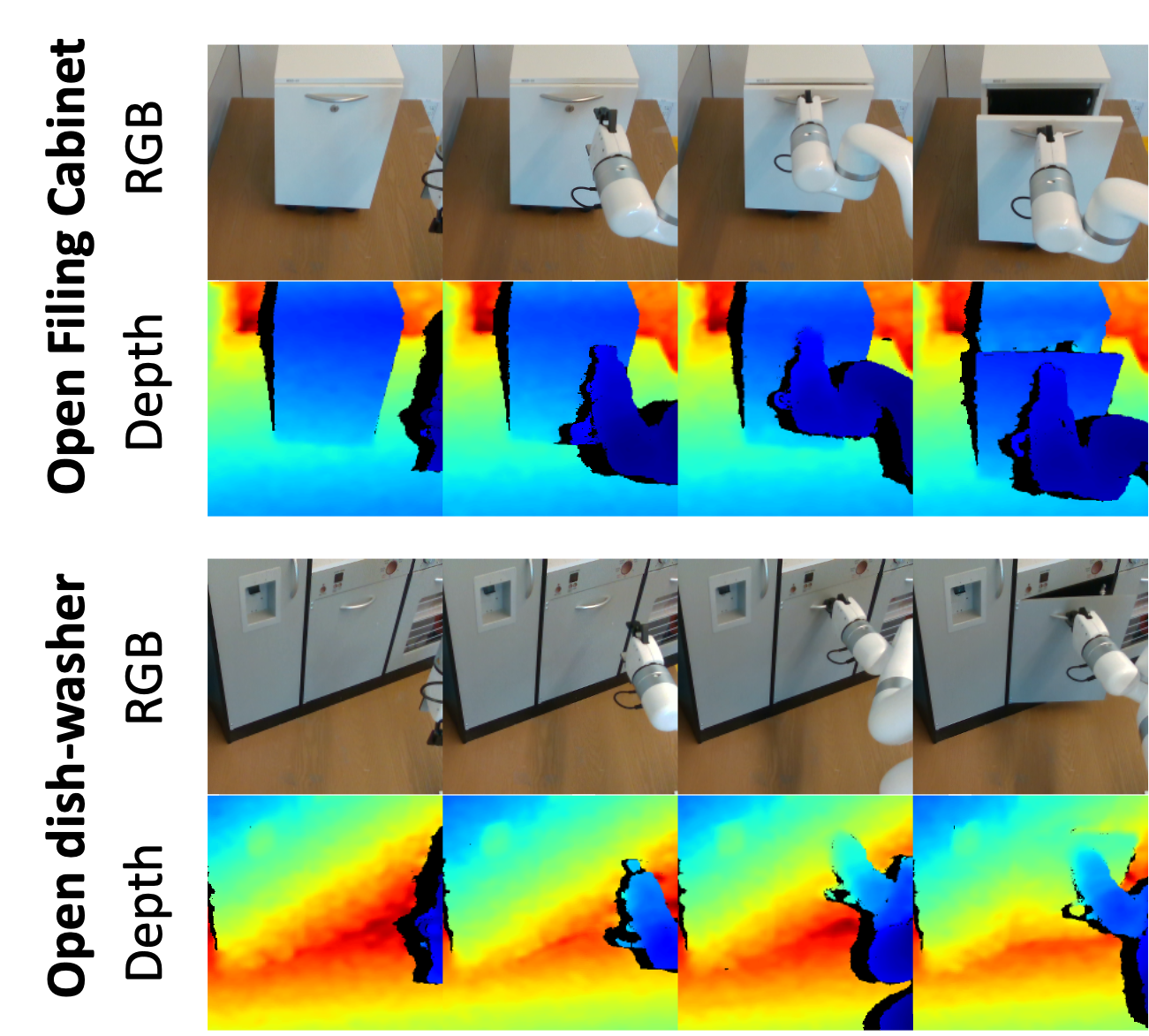}
\caption{Two of our real world tasks: (top) open filing cabinet, (bottom) open toy dishwasher. The agent only received 3 demonstrations. Each column for each tasks shows the RGB-D observations at $t=0$, $t=T/3$, $t=2T/3$ and $t=T$.}
\label{fig:real_tasks}
\end{figure}

C2F-ARM has already been qualitatively evaluated in the real world, and has demonstrated the ability to be trained efficiently from scratch~\cite{james2021coarse}. Our quantitative simulation experiments have already shown that our additions to C2F-ARM do not degrade performance. However, to further show this, we run an additional set of qualitative experiments in the real world, where we train on 2 real-world tasks from scratch, which can be seen in Figure \ref{fig:real_exp} and \ref{fig:real_tasks}. These tasks include opening a filing cabinet, and opening a toy dishwasher. At the beginning of each episode, the filing cabinet is moved randomly within the robot workspace; however, the kitchen (with the dishwasher) was not, due to its size. We train each of the tasks until the agent achieves 4 consecutive successes. The approximate time to train each task are $\sim 13$ minutes for open filing cabinet, and $\sim 10$ minutes for open dishwasher. We use the (low-cost) UFACTORY xArm 7, and a single RGB-D RealSense camera for all experiments. All tasks receive 3 demonstrations which are given through tele-op via HTC Vive. These qualitative results are best viewed via the videos on the project website.

\section{Conclusion} 
\label{sec:conclusion}

In this paper, we have presented \methName\ (\methAcro): an algorithm, which when given a next-best pose, learns to rank a set of paths generated from an array of path generating methods, including path planning, curve sampling, and a learned policy. The \methAcro\ is added as an extension to C2F-ARM, a highly efficient manipulation algorithm, and is able to achieve a wider set of tasks; in particular, tasks that require very specific motions that need to be inferred from both demonstrations and exploration data. 

There are a number of weaknesses that would make for exciting future work. Our path policy does not currently accept visual input, making our additional computational burden to C2F-ARM very low. Although this design decision is sufficient for all 16 RLBench tasks, it limits the role of the path policy to only complement the other two modules, rather than eventually replace them. For the path policy to be able to replace the other two modules, it would need access to vision. Therefore, an interesting future direction would be to investigate how we can incorporate vision into the ranking and path policy, without loss of sample efficiency. This is non-trivial, as C2F-ARM, by design, is agnostic to the number of cameras in the scene, therefore, the path policy would also need the same trait; perhaps by sharing features with the Q-attention. 
Although \methAcro\ has increased the repertoire of solvable tasks, C2F-ARM is still missing the closed-loop control that is needed for achieving tasks that have dynamic environments (e.g. moving target objects, moving obstacles, etc) or complex contact dynamics (e.g. peg-in-hole); exploring this in future work would further improve the system.
Finally, we have only explored ranking three path generation modules, however it would be interesting to explore if further performance gains could be had by introducing additional modules, such as other sample-based planners (e.g. SBL, KPIECE, etc), or optimization-based planners (e.g. TrajOpt, STOMP, etc).

\section*{Acknowledgments}
This work was supported by the Hong Kong Center for Logistics Robotics.

\bibliographystyle{unsrt}
\bibliography{main}

\end{document}